\documentclass[conference]{IEEEtran}
\IEEEoverridecommandlockouts

\usepackage{cite}
\usepackage{amsmath,amssymb,amsfonts}
\usepackage{algorithmic}
\usepackage{algorithm}
\usepackage{graphicx}
\usepackage{textcomp}
\usepackage{xcolor}
\def\BibTeX{{\rm B\kern-.05em{\sc i\kern-.025em b}\kern-.08em T\kern-.1667em\lower.7ex\hbox{E}\kern-.125emX}}

\usepackage{booktabs}    
\usepackage{array}       
\usepackage{tabularx}
\usepackage{multirow}
\usepackage{graphicx}  

\begin{document}

\title{
STAR: A Fast and Robust Rigid Registration Framework for Serial Histopathological Images\\
}

\author{
Zeyu Liu\textsuperscript{1}, 
Shengwei Ding\textsuperscript{2}\\[0.5em]
\textsuperscript{1} School of Biological Science and Medical Engineering, Beihang University, Beijing, China \\
\textsuperscript{2} School of Computer Science and Engineering, Beihang University, Beijing, China \\
zeyuliu@buaa.edu.cn
}
\maketitle

\begin{abstract}
Registration of serial whole-slide histopathological images (WSIs) is critical for enabling direct comparison across diverse stains and for preparing paired datasets in artificial intelligence (AI) workflows such as virtual staining and biomarker prediction. While existing methods often rely on complex deformable or deep learning approaches that are computationally intensive and difficult to reproduce, lightweight rigid frameworks—sufficient for many consecutive-section scenarios—remain underdeveloped. We introduce \textbf{STAR} (Serial Tissue Alignment for Rigid registration), a fast and robust open-source framework for multi-WSI alignment. STAR integrates stain-conditioned preprocessing with a hierarchical coarse-to-fine correlation strategy, adaptive kernel scaling, and built-in quality control, achieving reliable rigid registration across heterogeneous tissue types and staining protocols, including hematoxylin–eosin (H\&E), special histochemical stains (e.g., PAS, PASM, Masson’s), and immunohistochemical (IHC) markers (e.g., CD31, KI67). Evaluated on the ANHIR 2019 and ACROBAT 2022 datasets spanning multiple organs and scanning conditions, STAR consistently produced stable alignments within minutes per slide, demonstrating robustness to cross-stain variability and partial tissue overlap. Beyond benchmarks, we present case studies on H\&E–IHC alignment, construction of multi-IHC panels, and typical failure modes, underscoring both utility and limitations. Released as an open and lightweight tool, STAR provides a reproducible baseline that lowers the barrier for clinical adoption and enables large-scale paired data preparation for next-generation computational pathology. The code is available at \textit{https://github.com/Rowerliu/STAR}.
\end{abstract}


\section{Introduction}

Whole-slide image (WSI) registration has long been a cornerstone problem in computational pathology \cite{elhaminia2025traditional}. Accurate alignment across serial sections enables pathologists to directly compare tissue structures under different staining protocols, facilitating more comprehensive diagnostic assessment and interpretation \cite{lotz2015patch}. Beyond clinical use, registered multi-stain datasets provide paired information essential for emerging artificial intelligence (AI) applications, such as virtual staining, stain translation, and immunohistochemical (IHC) biomarker prediction from hematoxylin and eosin (H\&E) slides \cite{wodzinski2021deephistreg}. In both domains, robust registration acts as a bridge that links morphological and molecular perspectives, thereby advancing pathology toward more integrated and quantitative practices \cite{trahearn2017hyper}.

Despite decades of progress, most existing approaches to WSI registration emphasize deformable or learning-based models that prioritize high flexibility and local precision \cite{awan2018deep, wodzinski2019automatic}. While effective for handling tissue distortions, such methods are computationally intensive, often require large-scale training data, and are difficult to reproduce across laboratories \cite{shao2024raphia}. Moreover, the focus on complex algorithms has left a practical gap: there is no widely adopted, lightweight, and reusable rigid registration framework tailored to the needs of pathologists and AI researchers. In routine practice, rigid alignment often suffices for consecutive sections, where tissue distortion is relatively minor compared with inter-modality imaging problems \cite{muhlich2022stitching}. Yet the absence of a simple, standardized tool has hindered both clinical workflows and the preparation of large paired datasets for downstream analysis  \cite{roy2023deep}.

\begin{figure*}[htbp]
  \centering
    \includegraphics[width=0.9\linewidth, height=0.9\textheight, keepaspectratio]{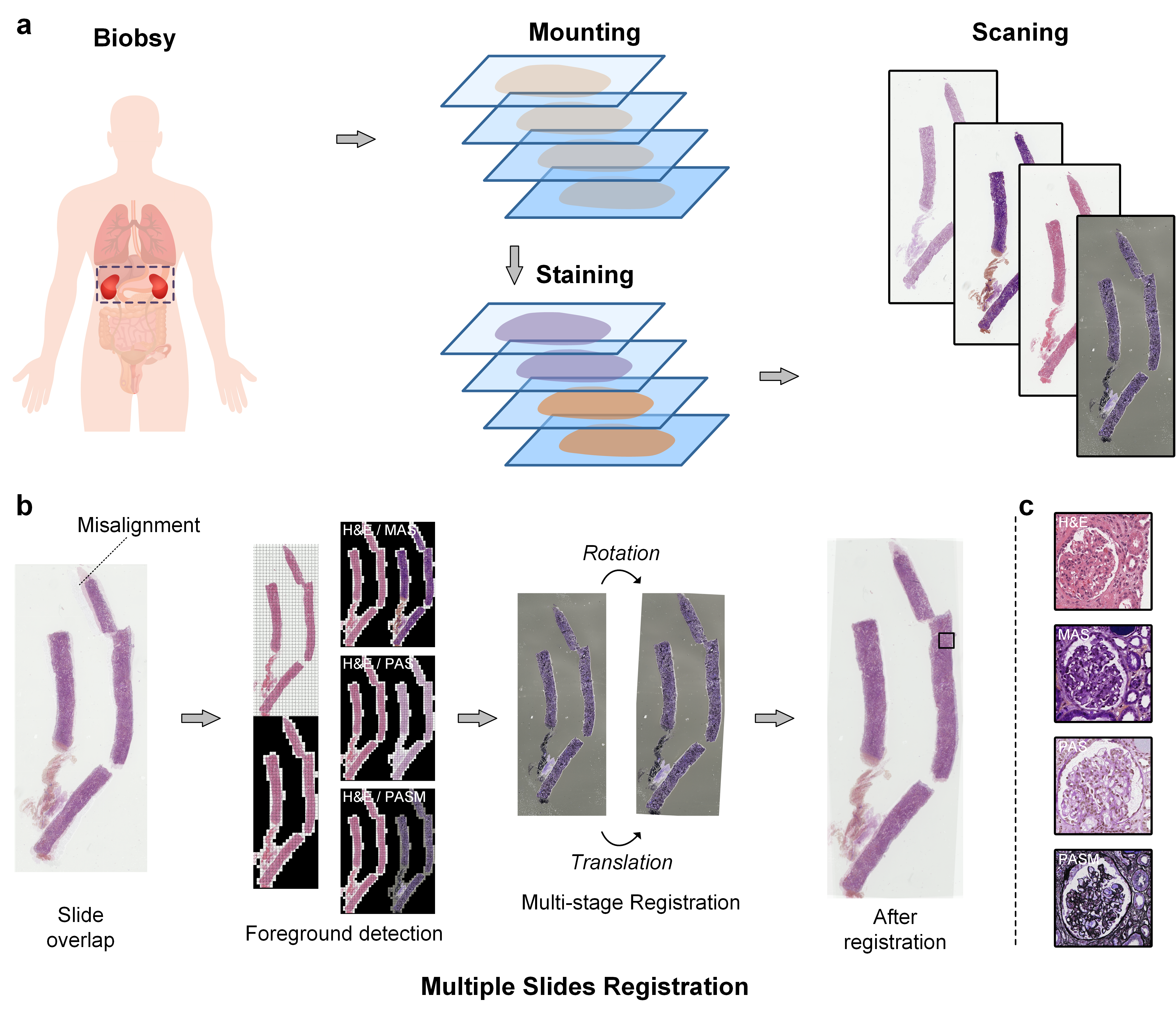}
  \caption{Overview of this study. \textbf{a}, Typical workflow of digital pathology, including biopsy, sectioning, staining, and scanning. \textbf{b}, The proposed STAR framework for serial rigid registration. \textbf{c}, Example of multi-stain images before and after alignment, showing the correspondence achieved by STAR.}
  \label{fig:fig_1}
\end{figure*}

To address this need, we propose \textbf{STAR} (Serial Tissue Alignment for rigid Registration), a fast and robust multi-WSI registration framework (Fig.~\ref{fig:fig_1}). Rather than introducing new deformation models, STAR focuses on the pragmatic requirements of usability, reproducibility, and scalability. Its contributions are as follows:
\begin{itemize}
    \item We introduce a reference-to-multi-target rigid registration pipeline designed for serial WSI alignment, emphasizing speed, robustness, and simplicity.
    \item The framework employs a multi-stage coarse-to-fine strategy, robust preprocessing against stain variability, and built-in quality control with fallback mechanisms.
    \item We release an open-source implementation optimized for batch processing of gigapixel slides, enabling practical adoption by both clinical and AI research communities.
    \item We demonstrate the clinical and computational relevance of STAR through use cases such as multi-stain comparison, virtual staining, and H\&E-based IHC prediction.
\end{itemize}

By filling the methodological gap between overly complex deformable approaches and the absence of usable rigid baselines, STAR seeks to establish a new community resource that is both practically useful and technically sound.

\section{Related Work}

\subsection{Rigid and Non-rigid WSI Registration}
WSI registration methods generally fall into rigid and non-rigid categories. Non-rigid approaches, including elastic models and modern deep learning frameworks, provide pixel-level alignment and can compensate for deformations, tears, or tissue folding \cite{hoque2022whole}. However, these methods typically demand high computational cost, rely on extensive parameter tuning or training data, and are rarely deployed in daily pathology workflows. Rigid registration, by contrast, is faster and less resource-intensive, and often sufficient for consecutive serial sections. Yet surprisingly few studies have systematically explored rigid WSI alignment as a standalone problem, and fewer still have released robust, open-source tools.

\subsection{Limitations of Existing Toolkits}
Several academic prototypes and commercial packages exist for histological registration, but most focus either on deformable models or narrow application scenarios \cite{chiaruttini2022open}. Their complexity often limits reproducibility, integration, and scalability, especially for large cohorts of multi-stain slides. Furthermore, benchmarking in the field has been dominated by accuracy metrics on small curated datasets, with less attention to usability and throughput. As a result, there remains a conspicuous lack of lightweight baselines that can serve as both practical tools for pathologists and standardized starting points for AI research.

\subsection{Positioning of STAR}
STAR is designed to fill this gap by offering a pragmatic, open-source, rigid registration framework optimized for multi-WSI alignment. Instead of competing with advanced deformable models, it provides a reliable baseline that prioritizes computational efficiency, cross-stain robustness, and reproducibility. In doing so, STAR complements existing work and enables both clinical workflows and algorithmic research to build upon a common foundation.

\section{Methods}

\subsection{Problem Formulation and System Overview}
Given a reference WSI $R$ (e.g., H\&E) and a set of $N$ target WSIs $\{T_i\}_{i=1}^N$ (e.g., IHC stains) acquired from serial sections, our objective is to rigidly align each $T_i$ to $R$. The alignment is parameterized by an in-plane rotation $\theta \in [0, 360)$ and a 2D translation $\mathbf{t}=(x,y)$. Let $\mathcal{R}_\theta(\cdot)$ denote rotation and $\mathcal{T}_{\mathbf{t}}(\cdot)$ denote translation. We seek
\begin{equation}
(\hat{\theta},\hat{\mathbf{t}}) = \arg\max_{\theta,\mathbf{t}} \; 
\langle \Phi_R,\; (\mathcal{T}_{\mathbf{t}} \circ \mathcal{R}_{\theta}) \, \Phi_T \rangle,
\end{equation}
where $\Phi_R$ and $\Phi_T$ are pre-processed grayscale representations of the reference and target images. In practice, this optimization is approximated by a coarse-to-fine search over rotated templates with correlation-based scoring. The STAR pipeline consists of: (i) reference foreground detection, (ii) stain-conditioned preprocessing, (iii) multi-stage rigid registration, and (iv) quality assessment and output generation.

\begin{algorithm}[t]
\caption{STAR: Multi-WSI Rigid Registration}
\label{alg:pr2_concise}
\begin{algorithmic}[1]
\REQUIRE Reference $R$, targets $\{T_i\}_{i=1}^{N}$; downsample $d{=}32$; angles $a_c{=}10^\circ$, $a_f{=}1^\circ$; strides $s_c{=}10$, $s_f{=}1$
\ENSURE Aligned crops \& masks; parameters $(\hat{x}_i,\hat{y}_i,\hat{\theta}_i)$

\STATE \textbf{Foreground on $R$}: build $d{\times}$ thumbnail; intensity gate (white$=230$, black$=20$), stride $64$; ResNet\textendash18 (256$\times$256) $\Rightarrow$ mask $M$; histogram-based ROI $\mathcal{B}$ (fallback by cumulative density).
\STATE \textbf{Ref preprocess}: crop $R$ to $\mathcal{B}$ at $d{\times}$; dilate $M$ ($7{\times}7$); equalize, invert; set $<50{\to}255$; affine rescale to $[0,255]$; zero background $\Rightarrow \Phi_R$.
\STATE \textbf{Rotation bank}: $\{\mathcal{R}_{\theta_k}\Phi_R\}_{\theta_k\in\{0,a_c,\ldots,360{-}a_c\}}$ on GPU.
\FOR{$i{=}1$ \TO $N$}
  \STATE \textbf{Tgt preprocess}: $d{\times}$ thumbnail of $T_i$; gray; set $<30{\to}255$; blur; invert $\Rightarrow \Phi_T$.
  \STATE \textbf{Coarse} ($a_c$, $s_c$): for each $\theta_k$, correlate $C_k=\Phi_T \star \mathcal{R}_{\theta_k}\Phi_R$ via CUDA \texttt{conv2d}; if kernel$>$input, bilinear downscale by $s{=}\min\{\frac{i_h-1}{k_h},\frac{i_w-1}{k_w}\}$; take $(x_c,y_c,\theta_c){=}\arg\max_{k,x,y} C_k[x,y]$.
  \STATE \textbf{Fine} ($a_f$, $s_f$): crop $\Phi_T$ around $(x_c,y_c)$ with $\pm50$ px $+$ half-template pad; search $\theta{\in}[\theta_c{-}10^\circ,\theta_c{+}10^\circ]$; get $(x_f,y_f,\Delta\theta)$; set $\hat{x}_i{=}x_f$, $\hat{y}_i{=}y_f$, $\hat{\theta}_i{=}\theta_c{+}\Delta\theta{-}10^\circ$.
  \STATE \textbf{Apply \& save}: rotate $T_i$ by $-\hat{\theta}_i$, translate to $(\hat{x}_i,\hat{y}_i)$; export thumbnail/native crops; make divider overlays; downsample $M$ to block mask for QA/patching.
\ENDFOR
\STATE \textbf{Optional}: tile native crops with \texttt{pyvips} (size/overlap user-defined), keep blocks where mask$=1$; enable integer-grid refinement on thumbnails, lift to native via crop-and-embed.
\end{algorithmic}
\end{algorithm}

\subsection{Reference Foreground Detection}
The registration pipeline begins with automated foreground detection to isolate tissue regions from background artifacts. We employ a pre-trained ResNet-18 architecture modified with a binary classification head to distinguish tissue from non-tissue regions. The network, denoted as $f_{\theta}: \mathbb{R}^{3 \times 256 \times 256} \rightarrow \{0,1\}$, operates on $256 \times 256$ pixel patches extracted at stride intervals of 64 pixels from down-sampled WSI representations.

For computational efficiency, input WSIs undergo initial downsampling by a factor of 32, transforming typical gigapixel images into manageable representations while preserving essential morphological features. The foreground detection process applies intensity-based pre-filtering, where pixels with grayscale values below 20 or above 230 are excluded to eliminate obvious background regions before network inference.

Following foreground prediction, we generate a binary mask $M \in \{0,1\}^{H \times W}$, where positive values indicate tissue presence. To establish the registration region of interest, we implement an adaptive bounding box extraction algorithm that identifies the minimal rectangular region containing the majority of foreground pixels. The bounding box extraction employs a density-based approach, analyzing pixel distributions along both spatial dimensions to determine optimal cropping boundaries while maintaining tissue integrity.

\subsection{Stain-Conditioned Preprocessing}
The preprocessing stage transforms raw WSI data into normalized representations suitable for template matching. This process differs substantially between reference and transform images to optimize registration performance across diverse staining protocols.

For reference images, typically H\&E sections, we apply a comprehensive normalization pipeline. The cropped tissue region undergoes histogram equalization using the \texttt{PIL.ImageOps.equalize} function to enhance contrast across the dynamic range. Subsequently, we apply intensity inversion to convert the conventional bright-background histological appearance to a dark-background representation more amenable to template matching. The preprocessing concludes with min-max normalization followed by a linear transformation:
\begin{equation}
    I_{norm} = \frac{(I_{eq} - I_{min})}{(I_{max} - I_{min})} \times 255,
\end{equation}
which represents the equalized image.

Transform image preprocessing follows a distinct pathway optimized for immunohistochemical and special stains. We first apply background thresholding to set pixels with intensity values below 30 to white (255), effectively removing artifacts and ensuring consistent background representation. The image then undergoes Gaussian blur filtering using \texttt{PIL.ImageFilter.BLUR} to reduce noise and enhance structural features. Finally, we apply intensity inversion through the transformation $I_{transform} = 255.0 - I_{input}$ to maintain consistency with the reference image preprocessing.

\subsection{Multi-stage Registration}

The core registration algorithm implements a hierarchical template matching approach utilizing rotational search across predefined angular ranges. The registration process consists of two sequential stages: coarse alignment with 10-degree angular increments, followed by fine alignment with 1-degree increments within a refined search window.

\subsubsection{Coarse Registration Stage}

The coarse registration stage generates a rotation matrix tensor $R \in \mathbb{R}^{N \times H \times W}$
 containing $N = 36$ rotated versions of the reference template, corresponding to angles $\theta_i = 10i$ degrees for $i \in \{0, 1, ..., 35\}$. Each rotation utilizes the OpenCV \texttt{cv2.getRotationMatrix2D} function with the template center as the rotation origin and unit scaling factor.

Template matching employs two-dimensional convolution between the transform image and each rotated template. For a given rotation angle $\theta_i$, we compute the convolution response:
\begin{equation}
    C_i = I_{transform} \star R_i,
\end{equation}
where $\star$ denotes correlation (implemented via \texttt{torch.nn.functional.conv2d} with stride parameter $s = 10$). The optimal coarse alignment parameters $(x_c, y_c, \theta_c)$ correspond to the maximum convolution response across all rotations:

\begin{equation}
    (x_c, y_c, \theta_c) = \arg\max_{i,x,y} C_i[x,y].
\end{equation}

\subsubsection{Fine Registration Stage}

The fine registration stage refines the coarse alignment within a restricted angular window $[\theta_c - 10°, \theta_c + 10°]$ using 1-degree increments. To improve computational efficiency, we crop the transform image to a region centered on the coarse alignment coordinates $(x_c, y_c)$ with padding of 50 pixels in each direction.

The fine-stage rotation matrix $R_{fine} \in \mathbb{R}^{21 \times H' \times W'}$ contains rotated templates at angles $\theta_j = \theta_c - 10 + j$ for $j \in \{0, 1, ..., 20\}$. The convolution process repeats with stride $s = 1$ to achieve pixel-level precision. The final registration parameters $(x_f, y_f, \theta_f)$ represent the optimal alignment transformation.

\subsubsection{Adaptive Kernel Scaling}
To handle cases where the template dimensions exceed the transform image size, particularly after cropping operations, we implement adaptive kernel scaling. When kernel dimensions $(k_h, k_w)$ exceed input dimensions $(i_h, i_w)$, we compute scaling factors:
\begin{equation}
s_h = \frac{i_h - 1}{k_h}, \quad s_w = \frac{i_w - 1}{k_w}.
\end{equation}

The kernel undergoes bilinear interpolation using $s = \min(s_h, s_w)$ to ensure compatibility while preserving template structure.

\subsection{Output Generation}

Output generation produces both thumbnail and full-resolution aligned image pairs. Thumbnail images undergo 16× downsampling for rapid visualization and quality assessment, while full-resolution outputs maintain original pixel spacing for downstream analysis. The system generates corresponding binary masks indicating tissue regions, enabling selective processing of relevant anatomical areas.

For multi-stain datasets, the framework supports batch processing with consistent reference frame alignment. Each transform image undergoes independent registration to the reference, followed by spatial transformation application to generate the final aligned dataset. The registration parameters $(x, y, \theta)$ are preserved for reproducibility and potential manual refinement when necessary.

The implementation handles edge cases including partial tissue overlap, rotation-induced boundary effects, and scaling variations through padding strategies and border handling protocols. These measures ensure robust performance across diverse tissue types and staining protocols commonly encountered in routine pathological practice.

\subsection{Patch Extraction and Optional Refinement}

Aligned native-resolution images are tiled into fixed-size patches with user-specified size and overlap (\texttt{pyvips} back-end). Blocks whose downsampled mask votes background are discarded, and filenames encode stain/slide/grid for downstream pairing. 

For difficult cases, we provide a human-in-the-loop refinement interface operating on thumbnails with divider grids; users input integer grid shifts that are then lifted to native-resolution crops. Implementation uses \texttt{pyvips} crop-and-embed to avoid memory overhead on large images.

\section{Experimental Settings}

\begin{figure*}[htbp]
  \centering
    \includegraphics[width=0.95\linewidth, height=0.95\textheight, keepaspectratio]{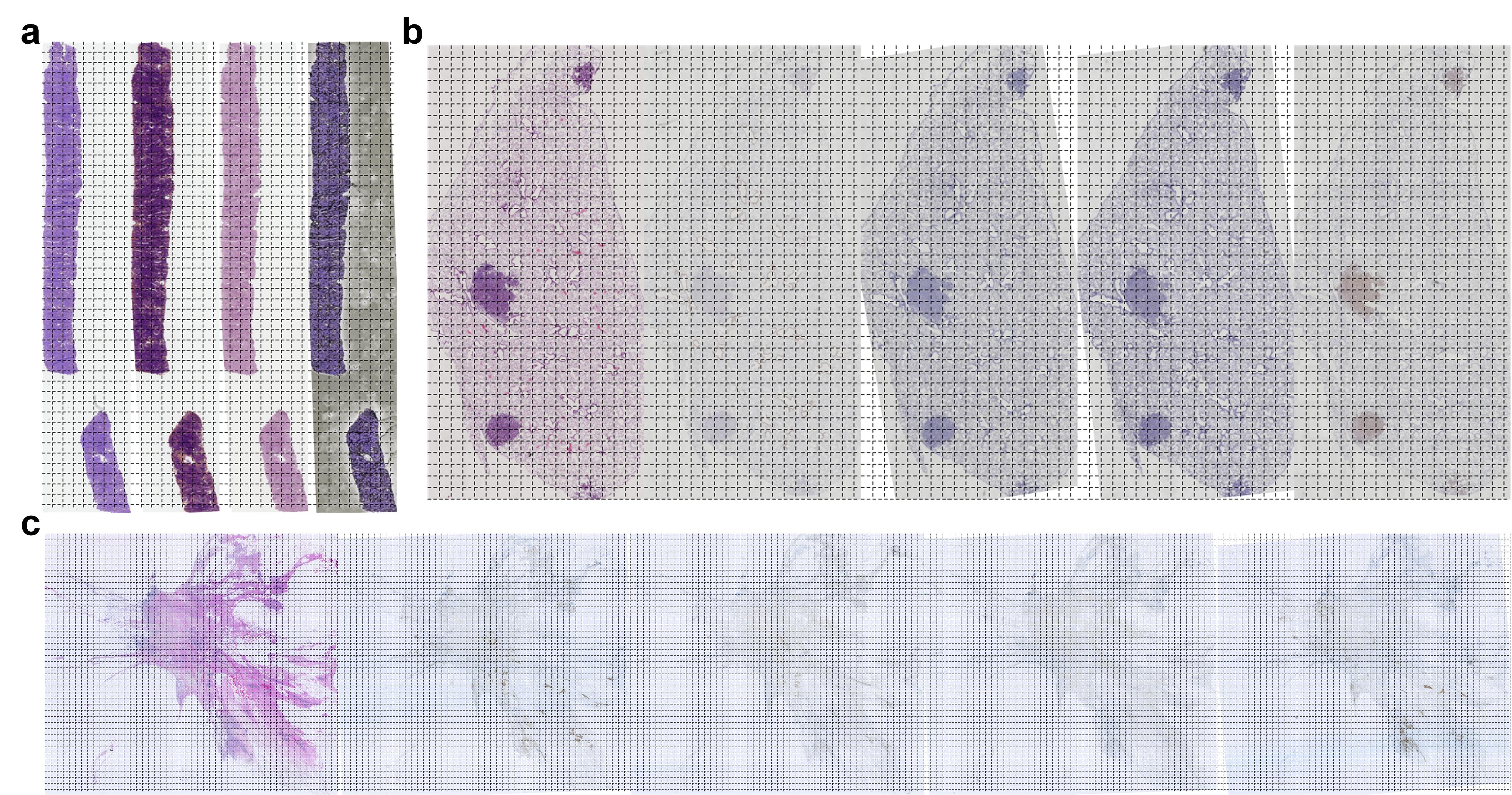}
  \caption{Representative registration outcomes across datasets. \textbf{a}, ANHIR (kidney) case including H\&E, MAS, PAS, and PASM stains. \textbf{b}, ANHIR (lung lobe) case including H\&E, CC10, CD31, KI67, and ProSPC. \textbf{c}, ACROBAT case including H\&E, ER, HER2, KI67, and PGR. STAR produces consistent alignment across diverse tissues and staining protocols.}
  \label{fig:fig_2}
\end{figure*}

\subsection{Datasets}
We evaluate STAR on two publicly available datasets widely used for histological registration tasks.

\textbf{ANHIR 2019.} The Automatic Non-rigid Histological Image Registration (ANHIR) challenge provides a large collection of serial histology sections across multiple organs and staining protocols \cite{borovecANHIRAutomaticNonRigid2020}. The dataset contains both rigid and non-rigid variations, with ground-truth landmarks available for quantitative evaluation. We use ANHIR primarily to assess the robustness of STAR against diverse tissue types and cross-stain differences. Although many ANHIR submissions emphasized non-rigid models, we focus on the rigid component to demonstrate the utility of STAR as a strong baseline.

\textbf{ACROBAT 2022.} The ACROBAT dataset was introduced as part of the 2022 histological image analysis challenge, focusing on multi-stain alignment across consecutive tissue sections \cite{weitzACROBAT2022Challenge2024}. It provides multi-institutional samples including H\&E, IHC, and special stains with carefully annotated correspondence. Compared with ANHIR, ACROBAT emphasizes clinical realism, containing slides with variable scanning conditions and partial tissue overlap. We employ ACROBAT to demonstrate the scalability of STAR in practical diagnostic scenarios and to validate its usability across heterogeneous laboratories.

Together, these datasets represent complementary challenges: ANHIR offers standardized landmark evaluation across multiple organs, while ACROBAT provides clinically realistic use cases with stain diversity and imaging variability. Evaluating on both ensures that STAR is tested not only for accuracy but also for generalizability and robustness.

\subsection{Implementation Details}
The software environment consisted of Python~3.10 with PyTorch~2.1, OpenCV~4.8, PIL, and \texttt{pyvips} for large-image handling. WSI files were accessed using \texttt{tifffile} and \texttt{OpenSlide}, with preprocessing and registration implemented as described in Section~III.

We set the default downsampling factor to $32\times$ for computational thumbnails, while visualization outputs were exported at $16\times$. The coarse registration stage used angular stride $a_c = 10^\circ$ and translation stride $s_c=10$ pixels, followed by a fine stage with $a_f=1^\circ$ and $s_f=1$ pixel. Background thresholds were fixed at 230 (white) and 20 (black) during foreground detection, and ResNet-18 was used with $256 \times 256$ patches at stride 64 for tissue classification. Batch size was set to 512 for efficient GPU utilization. For patch extraction, we adopted a default patch size of $256 \times 256$ with 20\% overlap, discarding blocks where the binary mask indicated background. 

The entire pipeline was designed for scalability: each target slide was processed independently against the reference, with foreground masks cached to disk to minimize redundancy. On the ANHIR dataset, rigid registration per slide pair required approximately 1--2 minutes, while ACROBAT processing times were similar despite greater stain variability, demonstrating the computational efficiency of STAR. All code and scripts are released as open source to ensure reproducibility and community adoption.

\section{Case Studies and Applications}

\subsection{Typical Use Cases}
One of the most common scenarios in computational pathology is the alignment of hematoxylin and eosin (H\&E) slides with corresponding special histochemical or immunohistochemical (IHC) stains. Such H\&E–IHC pairs are fundamental for supervised learning tasks, for example, predicting protein marker expression directly from morphology. STAR enables fast and reproducible alignment of such pairs, lowering the barrier to building large annotated datasets. Beyond pairwise registration, the framework also supports multi-chemical or multi-IHC panels, where multiple chem-stains (e.g., MAS, PAS, PASM) or immuno-stains (e.g., CD3, ER, KI67, HER2) are aligned to a single H\&E reference. This functionality provides a practical pathway for constructing multiplexed training sets without requiring expensive multi-modal imaging technologies.

\subsection{Qualitative Results on Different Tissue Types}
We illustrate the applicability of STAR across diverse tissue types and staining conditions using examples drawn from ANHIR and ACROBAT (Fig.~\ref{fig:fig_2}). In glandular tissues such as prostate and colon, rigid alignment was sufficient to capture major architectural correspondence despite natural variability in gland shapes. In tissues with complex microenvironments, such as lymph nodes or breast cancer sections, stain-specific preprocessing was crucial in enhancing contrast and suppressing background artifacts. Across both datasets, the resulting registered pairs facilitated side-by-side visualization, enabling pathologists to seamlessly compare nuclear morphology in H\&E with biomarker localization in IHC. Visual inspections confirmed that STAR provided stable alignment even in challenging conditions, including sections with partial tissue overlap or modest folding artifacts.

\subsection{Failure Cases and Limitations}
While STAR is designed for robustness and scalability, certain scenarios remain challenging. First, in slides with severe distortions, tears, or missing tissue regions, rigid alignment cannot fully compensate, and non-rigid models may still be necessary. Second, when consecutive sections differ substantially in cutting depth, fine-scale correspondence of cellular structures cannot be guaranteed even with accurate rigid alignment. Third, although preprocessing strategies improve cross-stain matching, extreme staining artifacts or scanner-specific noise can reduce correlation responses, occasionally leading to suboptimal alignment. Finally, while batch processing achieves scalability, real-time registration for interactive use remains outside the current scope. These limitations underscore that STAR is best positioned as a lightweight baseline and preprocessing tool, rather than a replacement for advanced deformable models.

\section{Community Impact and Future Work}

\subsection{Open-Source Ecosystem Contribution}
STAR is released as an open-source package with complete documentation, reproducible scripts, and modular APIs. By emphasizing simplicity and transparency, the framework lowers the entry barrier for pathologists, biomedical researchers, and computer vision scientists alike. In contrast to many previous registration tools that remain proprietary or are optimized for narrow research contexts, STAR is designed to integrate seamlessly into existing workflows. The open-source release is intended not only as a usable tool but also as a baseline reference implementation that other researchers can extend, compare against, and incorporate into broader pathology pipelines. We envision STAR serving as a community anchor for reproducible research in histological registration.

\subsection{Potential Applications in AI Training}
The availability of fast and reliable rigid registration opens new opportunities for training data preparation in computational pathology. Registered H\&E–IHC pairs can be directly used to supervise stain-to-stain translation networks, virtual staining methods, or predictive modeling of biomarker expression. Multi-IHC panels aligned to a common reference facilitate the construction of multiplexed learning tasks without specialized imaging technologies. Furthermore, robust batch alignment enables the assembly of large-scale paired datasets, which are critical for training foundation models in pathology. In this way, STAR not only addresses a practical gap in clinical workflows but also accelerates methodological advances in AI for pathology.

\subsection{Extension Possibilities}
Although STAR is focused on rigid registration, its modular architecture allows straightforward extension. Affine transformations could be integrated with minimal changes to the pipeline, providing flexibility for scaling or shearing effects. Non-rigid modules, whether based on classical deformation models or deep learning architectures, could be appended as optional refinement stages while preserving the lightweight rigid core. Beyond histology, the principles of STAR—multi-stage correlation, stain-robust preprocessing, and batch scalability—could also be applied to other biomedical imaging modalities, such as cytology smears or multiplex immunofluorescence. Future work will explore these extensions while maintaining the framework’s emphasis on usability and community accessibility.

\bibliographystyle{ieeetr}
\bibliography{main}

\end{document}